\documentclass[letterpaper]{article} % DO NOT CHANGE THIS
\usepackage{aaai23}  % DO NOT CHANGE THIS
\usepackage{times}  % DO NOT CHANGE THIS
\usepackage{helvet}  % DO NOT CHANGE THIS
\usepackage{courier}  % DO NOT CHANGE THIS
\usepackage[hyphens]{url}  % DO NOT CHANGE THIS
\usepackage{graphicx} % DO NOT CHANGE THIS
\usepackage{natbib}  % DO NOT CHANGE THIS AND DO NOT ADD ANY OPTIONS TO IT
\usepackage{caption} % DO NOT CHANGE THIS AND DO NOT ADD ANY OPTIONS TO IT
\usepackage{algorithm}
\usepackage{algorithmic}
\usepackage{amsmath}
\usepackage{amssymb}
\usepackage{multirow}
\usepackage{array}
\usepackage[table]{xcolor}
\usepackage{booktabs}

\usepackage{subfigure}
\makeatletter
\newcommand{\thickhline}{%
    \noalign {\ifnum 0=`}\fi \hrule height 1pt
    \futurelet \reserved@a \@xhline
}
\newcommand{\etal}{\textit{et al}.}
\newcommand{\ie}{\textit{i}.\textit{e}.}
\newcommand{\eg}{\textit{e}.\textit{g}.}
\newcommand{\ourmodel}{\textit{SSDA3D}}
\makeatother

\usepackage{newfloat}
\usepackage{listings}
\DeclareCaptionStyle{ruled}{labelfont=normalfont,labelsep=colon,strut=off} % DO NOT CHANGE THIS
\floatstyle{ruled}
\newfloat{listing}{tb}{lst}{}
\floatname{listing}{Listing}
\title{\textit{SSDA3D}: Semi-supervised Domain Adaptation for 3D Object Detection\\ from Point Cloud}
\author{
    Yan Wang \textsuperscript{\rm1}\thanks{Equal Contribution. Work done during internship at Inceptio.},
    Junbo Yin \textsuperscript{\rm1*},
    Wei Li \textsuperscript{\rm2}, 
    Pascal Frossard \textsuperscript{\rm3},
    Ruigang Yang \textsuperscript{\rm2}, 
    Jianbing Shen \textsuperscript{\rm4}\thanks{Corresponding author. Email: jianbingshen@um.edu.mo}\\
}
\affiliations{
    \textsuperscript{\rm 1}Beijing Institute of Technology \quad
    \textsuperscript{\rm 2}Inceptio \quad
    \textsuperscript{\rm 3}École Polytechnique Fédérale de Lausanne (EPFL) \\ 
    \textsuperscript{\rm 4}SKL-IOTSC, CIS, University of Macau

    yanwang@bit.edu.cn \quad yinjunbocn@gmail.com
}

\usepackage{bibentry}
\begin{document}

\maketitle

\begin{abstract}
LiDAR-based 3D object detection is an indispensable task in advanced autonomous driving systems. Though impressive detection results have been achieved by superior 3D detectors, they suffer from significant performance degeneration when facing unseen domains, such as different LiDAR configurations, different cities, and weather conditions. 
%%%%
The mainstream approaches tend to solve these challenges by leveraging unsupervised domain adaptation (UDA) techniques. However, these UDA solutions just yield unsatisfactory 3D detection results when there is a severe domain shift, \eg, from Waymo (64-beam) to nuScenes (32-beam). 
% To this end, 
% we present a novel and more realistic adaptation setting in this paper, \ie, 
To address this,
we present a novel \textbf{S}emi-\textbf{S}upervised \textbf{D}omain \textbf{A}daptation method for \textbf{3D} object detection (\ourmodel), where only a few labeled target data is available, yet can significantly improve the adaptation performance. 
%%%%%%%%
In particular, our \ourmodel~includes an Inter-domain Adaptation stage and an Intra-domain Generalization stage.
In the first stage, an {Inter-domain} Point-CutMix module is presented to efficiently align the point cloud distribution across domains. The Point-CutMix generates mixed samples of an intermediate domain, thus encouraging to learn domain-invariant knowledge. Then, in the second stage, we further enhance the model for better generalization on the unlabeled target set. This is achieved by exploring {Intra-domain} Point-MixUp in semi-supervised learning, which essentially regularizes the pseudo label distribution. Experiments from Waymo to nuScenes show that, with only 10$\%$ labeled target data, our \ourmodel~can surpass the fully-supervised oracle model with 100$\%$ target label. Our code is available at  https://github.com/yinjunbo/SSDA3D.

\end{abstract}

\section{Introduction}

%----------- Figure 1 ----------- %
\begin{figure}[h]
\begin{center}
\includegraphics[width=0.45\textwidth]{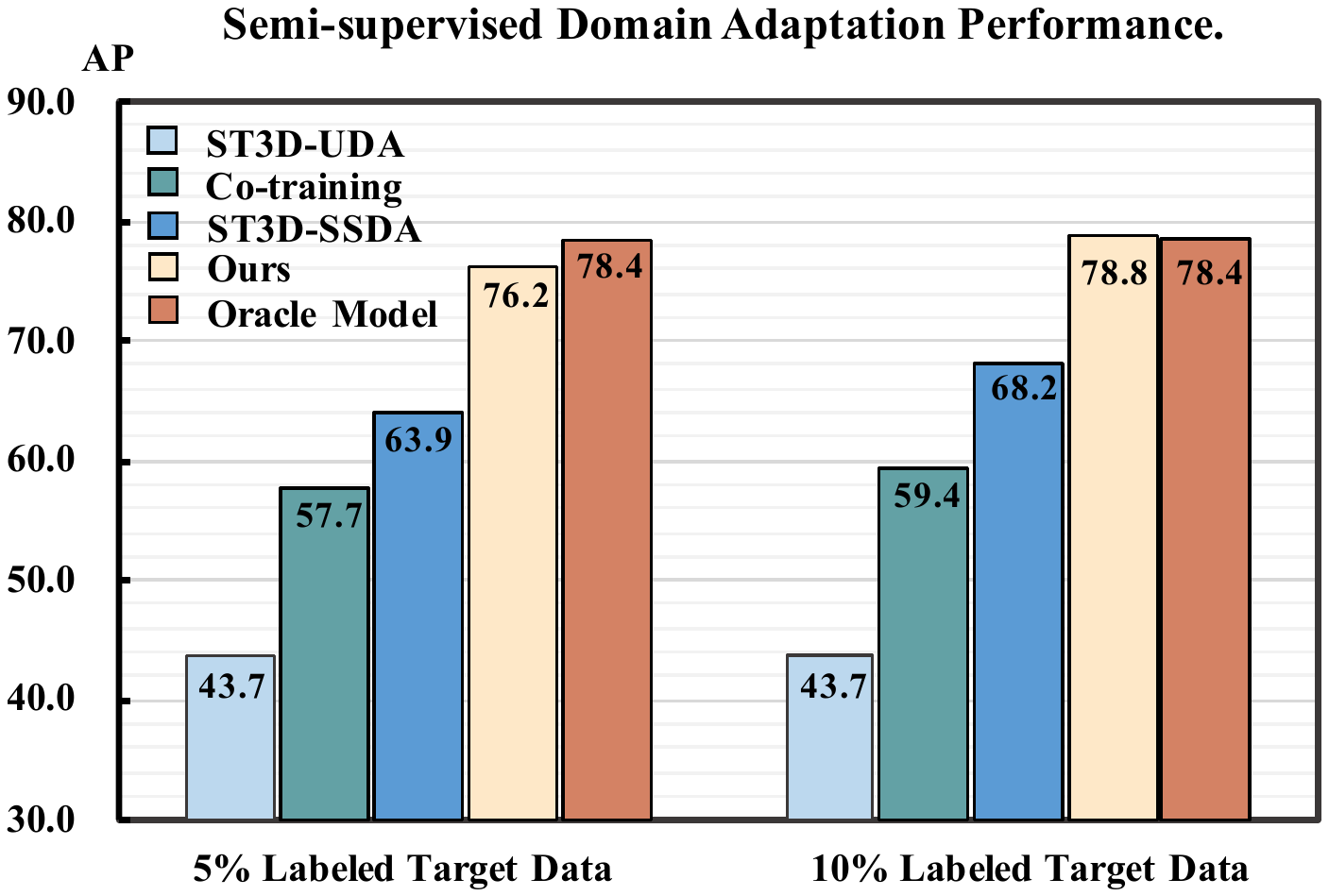}
\end{center}
% \vspace{-4mm}
\caption{\small \textbf{An example of semi-supervised domain adaptation on Waymo$\rightarrow$nuScenes (5\% and 10\%~label)} based on the CenterPoint~\cite{CENTERPOINT_2021} detector. Our~\ourmodel~significantly surpasses advanced methods such as {Co-training} (\ie, jointly train the labeled source and target data) and {ST3D-SSDA} (\ie, adapt ST3D~\cite{yang2021st3d} with the target label). Moreover, with only 10\% target label, we also exceed the \textit{Oracle} model that is trained with 100\% target label.}
%\vspace{-6mm}
\label{fig:ssda}
\end{figure}

3D object detection from LiDAR point cloud has been a central task in applications such as robotics and autonomous driving. It aims to localize and classify the obstacles in the 3D space accurately and timely, where the obstacles are often formulated as orientated 3D bounding boxes. In the past few years, substantial progress has been made in 3D object detection thanks to the well-developed 3D neural networks such as VoxelNet~\cite{VoxelNet_2018_Zhou} and PointNet~\cite{PointNet++_2017}. The advanced 3D object detectors~\cite{Second_2018Yan, CENTERPOINT_2021, PVRCNN_2020_Shi, xu2021fusionpainting} can gain promising detection results when given massive well-labeled training samples. Unfortunately, these 3D detectors may incur dramatic performance drops when they are deployed in unseen environments. 
% For example, the 3D point cloud collected from different geographical locations may present various street styles and car sizes~\cite{wang2020train}. 
%%%%
% The LiDAR sensor is also {unstable} when working in adverse weather conditions like rain~\cite{byeon2020analysis} or snow~\cite{hahner2022lidar}. 
% Further, different LiDAR sensor types~\cite{yang2021st3d} and the placement variance of the sensor~\cite{hu2022investigating} remarkably influence the 3D detection performance. 
The domain shift~\cite{sun2016return,tzeng2017adversarial} indicates that there is an obvious distribution discrepancy between the training data (source domain) and test data (target domain). One solution to overcome this challenge is to further manually annotate the target domain data. 
However, it requires tremendous effort and high cost to annotate in 3D point cloud, and it is also not feasible to cover all the cases in the wild. Therefore, how to tackle the domain shift problem in 3D object detection remains an open challenge.

A recent trend is to exploit unsupervised domain adaptation (UDA) algorithms to mitigate the domain gap in 3D object detection, where SRDAN~\cite{zhang2021srdan}, SN~\cite{DBLP:conf/cvpr/WangCYLHCWC20} and ST3D~\cite{yang2021st3d} are the prior works. 
%SRDAN performs domain alignment between the source and target data by applying adversarial learning on different feature levels. SN leverages the target statistics such as 3D object sizes to adjust the source domain data. ST3D further adapts self-training on 3D UDA by assigning pseudo labels on the target domain. 
%These approaches largely close the gaps in domains with similar LiDAR configuration,
Though impressive domain adaptation ability has been demonstrated between domains with similar LiDAR configuration, \eg, from Waymo~\cite{WAYMO_2020Sun} to KITTI~\cite{geiger2012we}, where both are based on 64-beam Velodyne LiDAR, none of them is capable of tackling the domain gap derived from various LiDAR configurations. For example, on the Waymo (64-beam)$\rightarrow$nuScenes (32-beam)~\cite{caesar2019nuscenes} adaptation setting, ST3D only gains 2 to 3 points improvement over the \textit{source only} model (\ie, trained with only the Waymo data), even with the assistance of target statistics information. Nevertheless, it is often the case that a self-driving car will upgrade its LiDAR sensor to a new type, and it seems that the well-labeled source data from the old LiDAR sensor is not able to contribute much with current UDA algorithms, which leads to a huge waste. 

The above observations motivate us to devise a more realistic setting, \ie, semi-supervised domain adaptation (SSDA). Compared to UDA, SSDA smartly uses only a small set of labeled target data, together with the large-scale labeled source data. It can greatly close the gap to the fully-supervised \textit{Oracle} model, meanwhile maintaining a rather lower annotation cost that is readily acceptable in the practical application. An example of SSDA is shown in Figure~\ref{fig:ssda}, where we aim to adapt Waymo to nuScenes and only a few labeled nuScenes data is available. An intuitive way for addressing this is to jointly train the labeled source and target data, which is called {Co-training}. 
%According to Fig.~\ref{fig:ssda}, \textit{Co-training} even produces worse results than \textit{Target Sup.}, \ie, train with only the labeled part of target. This can be attributed to the inappropriate domain mixing, where the large-scale source domain dominates the learning process. 
This obtains suboptimal performance due to the inappropriate domain mixing, where the large-scale source domain dominates the learning process. 
An alternative solution is to adapt UDA techniques to SSDA. We enhance ST3D with the labeled part of nuScenes and name it {ST3D-SSDA}. Despite the better performance than {Co-training}, it is still far behind the fully-supervised \textit{Oracle} that is trained with   the full target label. 

In this work, we propose a new framework to tackle SSDA in 3D object detection, named~\textit{SSDA3D}. The core idea of~\ourmodel~is to reduce the Inter-domain discrepancy (\eg, labeled source$\rightarrow$ labeled target) as well as enhance the intra-domain generalization (\eg, labeled target$\rightarrow$unlabeled target). This is realized by solving domain adaptation task and semi-supervised learning (SSL) task in a unified framework by two-stage learning, 
which includes an \textbf{Inter-domain Adaptation Stage} and an \textbf{{Intra-domain Generalization Stage}}.
% We aim to achieve domain adaptation in the first stage with labeled source and target data and improve the generalization in the second stage which we call \textbf{Inter-domain Adaptation Learning Stage} and \textbf{Intra-domain Generalization Learning Stage} respectively. 
Concretely, in the first stage, we aim to generate mixed samples from the labeled source and target data, which acts as intermediate domain data that can mitigate the domain bias and learn domain-invariant representation. This is achieved by an Inter-domain Point-CutMix module, \eg, we randomly remove a local point cloud region in the source sample, and then replace it with another region from the target sample to close the distribution discrepancy. 
Regions from different point cloud range present various patterns, \eg, the points become more sparse in distant regions. Thus, a constraint is enforced that the inpainting regions should be from the same range to preserve the geometrical nature. The model in the first stage has significantly boosted the performance thanks to the efficient domain mixing. In the second stage, we aim to further strengthen the intra-domain generalization on the unlabeled target data. In particular, the pseudo-labeling technique is typically used in SSL for learning from the unlabeled data, while we find that inaccurately pseudo-labeled samples essentially account for a higher percentage, which will dominate the learning and undermine the performance. To this end, an Intra-domain Point-MixUp module is introduced by globally mixing the labeled scenes and the pseudo-labeled scenes. In this way, the mixed data and labels can take the role of regularizer to implicitly improve the learning process.

To the best of our knowledge,~\textit{SSDA3D} is the first effort for semi-supervised domain adaptation in the context of 3D object detection. This is achieved by a novel framework that jointly addresses inter-domain adaptation and intra-domain generalization. The proposed Inter-domain Point-CutMix module largely reduces the domain discrepancy, while the Intra-domain Point-MixUp module essentially regularizes the pseudo label distribution of unlabeled target data. Our model is evaluated in a challenging domain adaptation setting, \eg, Waymo$\rightarrow$nuScenes. As shown in Figure~\ref{fig:ssda}, with only 10\% labeled target data, we achieve competitive performance to the fully-supervised \textit{Oracle} model. It turns out that our method can save almost 90\% annotation cost for cross-domain 3D object detection. 

% The contributions of this work are as follows:  
% \begin{itemize}
% \item A new task, SSDA for 3D object detection, is presented in this work. We make {the first effort} for handling this task by proposing~\textit{SSDA3D}, a novel framework that jointly addresses inter-domain adaptation and intra-domain generalization. 
% % This will encourage further research and application of this topic.
% \item An Inter-domain Point-CutMix module is proposed for learning from intermediate domain and abstracting domain-invariant knowledge from both source and target data. This largely reduces the domain discrepancy.
% \item We also introduce an Intra-domain Point-MixUp module to improve the semi-supervised learning in the target domain. The mixed point cloud and label essentially regularize the pseudo label distribution of unlabeled data.
% \item We evaluate our model on a challenging domain adaptation setting, \eg, Waymo$\rightarrow$nuScenes, and~\ourmodel~significantly outperforms previous approaches. Moreover, as shown in Fig.~\ref{fig:ssda}, with only 10\% labeled target data, we achieve competitive performance to the fully-supervised \textit{Oracle} model. It turns out that our method can save almost 90\% annotation cost for 3D object detection. 
% \end{itemize}

%----------- Figure 2 ----------- %
\begin{figure*}
\includegraphics[width=0.85\linewidth]{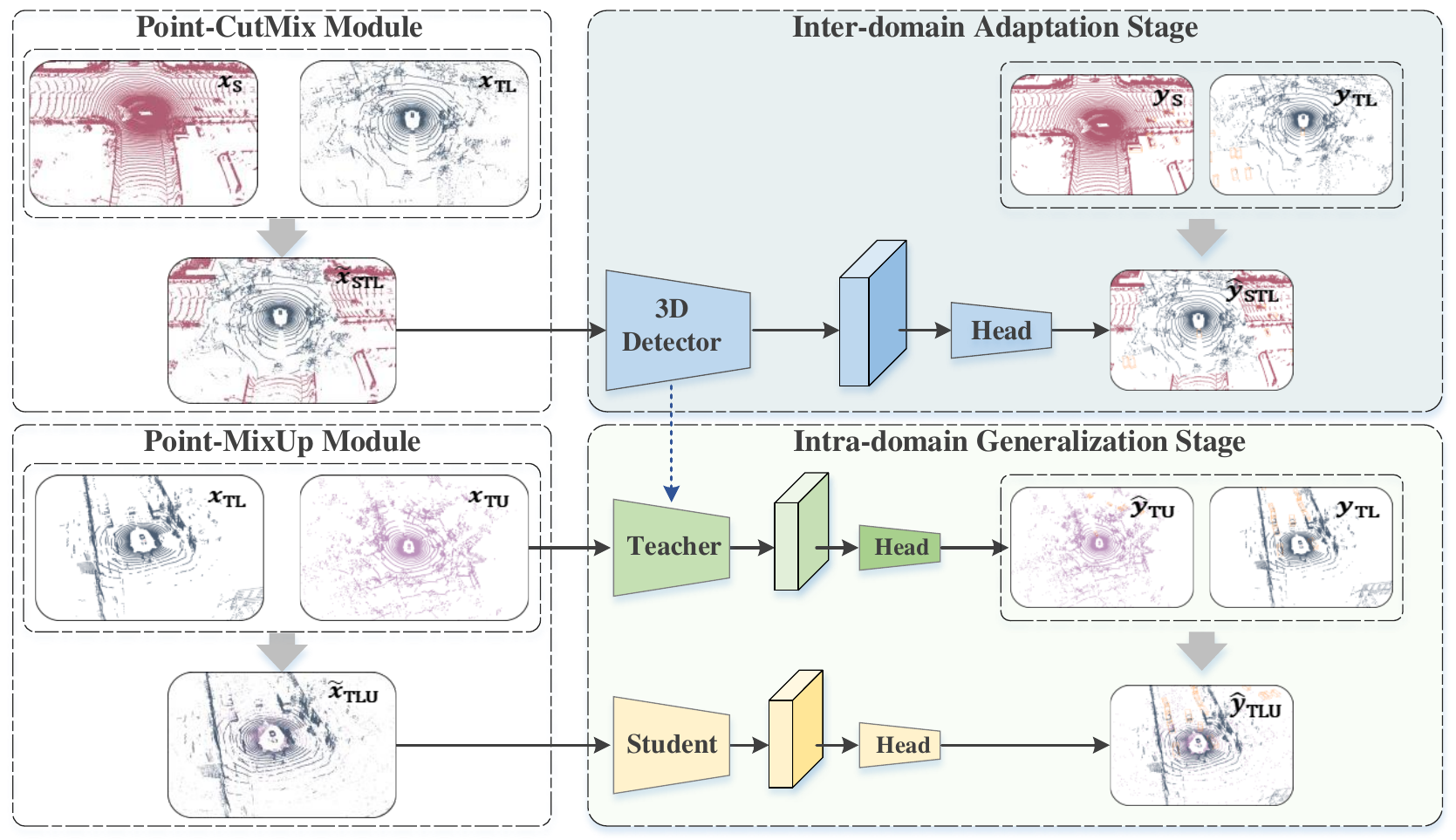}
\centering
% \vspace{-2mm}
\caption{\small \textbf{The overall framework of our~\ourmodel.} It comprises an Inter-domain Adaptation stage and an Intra-domain Generalization stage. The Point-CutMix in the former stage generates region-wise mixed samples from source and target to address the domain discrepancy. The Point-MixUp in the latter stage further constructs scene-wise mixed samples for improving the learning on the unlabeled set.}
\label{pipeline.pdf}
\end{figure*}

\section{Related Works}
\noindent\textbf{3D Object Detection from LiDAR Point Cloud.}
LiDAR-based 3D object detection has received great attention due to the rapid development of autonomous driving. The mainstream 3D object detection approaches can be categorized into three groups, \eg, point-based methods~\cite{PointNet++_2017,PointRCNN_2019SHI,yang20203dssd,zhang2022not,zhou2020joint,chen2022sasa}, voxel-based methods~\cite{PointPillars_2018LANG,Second_2018Yan,CENTERPOINT_2021,zheng2021cia} and hybrid methods with joint point-voxel representation~\cite{PVRCNN_2020_Shi,mao2021pyramid,deng2021voxel}. 
% Point-based methods~\cite{PointNet++_2017,PointRCNN_2019SHI,yang20203dssd,zhang2022not,zhou2020joint,chen2022sasa} describe the 3d space via the raw-point representation, which typically use backbones like PointNet++~\cite{PointNet++_2017} to extract features. These methods are often restricted to the inference speed when the point scale becomes large.
Point or hybrid approaches are often restricted to the inference speed when the point scale becomes large.
Voxel-based approaches are more prevalent in practical applications, since they can engage trade-offs between accuracy and speed. They usually discretize the 3D space into regular voxels, and then apply sparse convolutional networks~\cite{graham20183d,chen2022focal} for abstracting 3D features. 
% Another trend is to combine the advantages of point-based and voxel-based methods and exploits the hybrid representation~\cite{PVRCNN_2020_Shi,mao2021pyramid,deng2021voxel}. To achieve better results, they often require an additional stage for box refinement. 
Apart from the single-frame 3D object detection paradigm, there are also some attempts to leverage the temporal point cloud information~\cite{yin2021graph, chen2022mppnet}.
However, all these 3D detectors work well when given large-scale precisely annotated point cloud samples, but are incapable of handling unseen domains without sufficient annotations. In this paper, we investigate the semi-supervised domain adaptation in 3D object detection, which effectively improves the domain generalization of leading 3D detectors.\\

\noindent\textbf{Semi-supervised Learning in 3D Object Detection.} To achieve promising detection performance, prevalent 3D object detectors resort to large-scale 3D annotations. However, collecting such annotations is extremely time-consuming and expensive. This motivates researchers to apply semi-supervised learning(SSL) methods~\cite{tang2019transferable, yin2022proposalcontrast} to cut the expense.
% Tang~\etal~ \cite{tang2019transferable} proposes to train 3D object detector for new classes from only 2D bounding box labels of these new classes while transferring information from existing 3D bounding box labels at the same time. thus saving the annotation cost. 
In order to boost the performance with limited annotations, SESS \cite{SESS_2020} proposes a self-ensembling framework to enhance the generalization ability on unlabeled data. After that, 3DIoUMatch \cite{3DIoUMatch_2021} proposes an IoU-based filtering mechanism to improve the quality of the pseudo labels. Later, ProficientTeachers \cite{yin2022semi} leverages a spatial-temporal ensemble module and a clustering-based box voting module to further refine the pseudo labels. There are also some works exploring weakly semi-supervised 3D object detection to save annotation cost. Meng~\etal~\cite{meng2021towards} propose to learn from a few weakly annotated point cloud samples as well as some precisely annotated instances. It achieves 97\% performance to the oracle model by a two-stage network architecture. These methods mainly focus on intra-domain 3D SSL, while ours aims to address cross-domain 3D SSL which needs to transfer knowledge from a well-annotated source domain to a few-annotated target one and faces more challenges.

\noindent\textbf{Domain Adaptation in Point Cloud.}
Recent years have witnessed an increased interest in domain adaptation in 3D point cloud \cite{achituve2021self, xiao2022transfer, peng2020point, jiang2021lidarnet}, which mainly focuses on unsupervised domain adaptation (UDA). For UDA in 3D semantic segmentation~\cite{xu2021spg}, SqueezeSegV2~\cite{wu2019squeezesegv2} proposes several techniques  to adapt simulated data to the real world, such as intensity rendering, geodesic alignment and layer-wise calibration between domains. ePointDA~\cite{zhao2021epointda} also handles the simulation-to-real setting. It employs CycleGAN for dropout noise rendering and performs statistics-invariant feature alignment to close the gap. To address domain discrepancies caused by different sensors, Langer~\etal~\cite{langer2020domain} generate semi-synthetic LiDAR scans to simulate the target LiDAR sensor, and also align outputs between different domains via the geodesic loss.  Later, SVCN~\cite{DBLP:conf/cvpr/YiGF21} designs a point cloud completion network to recover both the source and target points to a canonical domain (\eg, the complete point cloud), and then performs segmentation on this canonical domain. Several studies have also explored UDA in 3D object detection~\cite{saleh2019domain}. SRDAN~\cite{zhang2021srdan} performs domain alignment by applying adversarial learning on different feature levels. SN~\cite{DBLP:conf/cvpr/WangCYLHCWC20} leverages the target statistics such as 3D object sizes to adjust the source domain data. ST3D~\cite{yang2021st3d} further adapts self-training on 3D UDA by assigning pseudo labels on the target domain. The experiments indicate that it is extremely challenging for these UDA approaches to catch the oracle performance, especially when the domain gap is large (e.g., cross LiDAR configurations). By contrast, our~\ourmodel~gains much better performance, meanwhile saving 90\% annotation cost. \\
\\
\noindent\textbf{Mixing Augmentation Strategies.}
Data augmentation~\cite{cubuk2020randaugment,cubuk2019autoaugment}, which has been proved an indispensable component in training deep neural networks, aims to generate new training samples lie in the vicinity distribution of the original samples. Different from the traditional augmentation strategies such as random crop, flip, rotation and scaling, MixUp~\cite{zhang2017mixup} constructs new training samples by linearly combing two samples drawn from the training set. CutMix~\cite{yun2019cutmix} also aims to combine two different training samples. Instead of directly mixing the two samples globally, it replaces the local image patch of one sample with another patch from a different sample. In essence, both MixUp and CutMix are designed for the image classification task. In this work, we adapt these techniques to 3D object detection in point cloud, which is an early effort that exploits mixing augmentation in 3D object detection. The analysis in Sec.~\ref{subsec:framework} reveals their capacity in addressing inter-domain discrepancy and intra-domain generalization.

\section{Our Semi-supervised Domain Adaptation \\Learning Approach}

% pipeline

%\subsection{Problem Statement}
%For SSDA in 3D Object detection, we take as input three types of point cloud data from two different domains, \eg, the large-scale \textbf{labeled source} data, a small set of \textbf{labeled target} data and the large-scale \textbf{unlabeled target} data. Let ${D_S}$ = ${ \left\{ ( x_S^i, y_S^i ) \right\} }_{i=1}^{N_S}$ be the ${N_S}$ labeled source point cloud samples, and ${D_{TL}}$ = ${ \left\{ ( x_{TL}^i, y_{TL}^i ) \right\} }_{i=1}^{N_{TL}}$ be the ${N_{TL}}$ labeled target point cloud samples, and ${D_{TU}}$ = ${ \left\{ ( x_{TU}^i ) \right\} }_{i=1}^{N_{TU}}$ be the ${N_{TU}}$ unlabeled target point cloud samples. Here, $x_S^i$ and $y_S^i$ represent the \textit{i}-th labeled source point cloud and its corresponding label. Similarly, $x_{TL}^i$ and $y_{TL}^i$ represent the \textit{i}-th labeled target point cloud and its corresponding label, and $x_{TU}^i$ represents the \textit{i}-th unlabeled target point cloud. Our purpose is to develop a method to effectively leverage the data mentioned above to boost the detection performance on target point cloud dataset.
Previous approaches tend to tackle the point cloud domain adaptation by UDA, while they are incapable of handling large domain discrepancies. For example, Waymo~\cite{WAYMO_2020Sun} utilizes a 64-beam LiDAR with beam angle $[-18^\circ, 2^\circ]$, while nuScenes~\cite{caesar2019nuscenes} adopts a 32-beam LiDAR with beam angle $[-30^\circ, 10^\circ]$. Hence, their point cloud patterns are of great difference. The two datasets are also collected from different geographical locations, leading to environment domain gaps such as road styles and object scales. A prior work, ST3D~\cite{yang2021st3d}, reports an improvement of 3.0 AP with the help of nuScenes object statistics, which is far from satisfactory for practical deployment in self-driving vehicles. Thus, a better way is to explore SSDA for addressing this challenge.

SSDA has limited access to a small set of labeled target samples, thus it is of great importance to fully leverage these training samples. A naive method is to directly train the labeled source and target data simultaneously, which is called {Co-training}. However, this turns out to be ineffective and even produces worse results than training the labeled target data only. This can be attributed to inappropriate domain mixing, \eg, the large-scale source domain dominates the learning process. Another solution is to extend the state-of-the-art UDA framework to SSDA. A typical method, ST3D~\cite{yang2021st3d}, first trains on the source domain with random object scaling for reducing the domain gap, and then generates pseudo labels for the unlabeled target data. We adapt ST3D to {ST3D-SSDA}, where both stages are equipped with the labeled target data. This significantly improves the performance compared to the UDA solutions (plz refer to Figure~\ref{fig:ssda}), proving the importance of SSDA.

In this paper, we develop a new SSDA framework, \ourmodel, for 3D object detection. 
%that can effectively transfer knowledge from $D_S$ to $D_{TL}$, as well as can generalize to the unlabeled set $D_{TU}$. 
In Sec.~\ref{subsec:framework}, we detail the overall learning pipeline of our~\ourmodel.
%which is composed of an Inter-domain Adaptation stage and an Intra-domain Generalization stage. 
Then, in Sec.~\ref{subsec:cutmix}, a Point-CutMix module is advocated to address domain 
adaptation problem between source and target. Afterwards, a Point-MixUp is presented in Sec.~\ref{subsec:mixup} to further improve the learning on the unlabeled target data.

\subsection{Two-stage Learning for SSDA}
\label{subsec:framework} 
For SSDA in 3D Object detection, we take as input three types of point cloud data from two different domains, \eg, the {labeled source} data  ${D_\text{S}}={ \left\{ ( x_\text{S}^i, y_\text{S}^i ) \right\} }_{\text{i=1}}^{N_\text{S}}$, the {labeled target} data ${D_{\text{TL}}}={ \left\{ ( x_{\text{TL}}^i, y_{\text{TL}}^i ) \right\} }_{\text{i=1}}^{N_{\text{TL}}}$ and the {unlabeled target} data ${D_{\text{TU}}}={ \left\{  x_{\text{TU}}^i \right\} }_{i=1}^{N_{\text{TU}}}$, where 
 ${N_{\text{S}}}$, ${N_{\text{TL}}}$ and $N_{\text{TU}}$ are the number of data samples. A crucial characteristic of SSDA is to leverage the well-labeled source data to save annotations of the target, thus both ${N_{\text{S}}}$ and  $N_{\text{TU}}$ are much bigger than ${N_{\text{TL}}}$, \ie, ${N_{\text{S}}}\gg {N_{\text{TL}}}$ and ${N_{\text{TU}}}\gg{N_{\text{TL}}}$. Besides, $x^i$ and $y^i$ represent the \textit{i}-th point cloud sample and corresponding 3D detection label, respectively.

Since the SSDA task requires to resolve both cross-domain and intra-domain generalization, we thus advocate a two-stage learning strategy. As illustrated in Figure \ref{pipeline.pdf}, the first stage is the Inter-domain Adaptation stage, which is used to tackle the domain discrepancy between the source ${D_{\text{S}}}$ and labeled target ${D_{\text{TL}}}$. To achieve this, mixed point cloud samples ${{D}_{\text{STL}}}={ \left\{ ( \tilde{x}_{\text{STL}}^i, \tilde{y}_{\text{STL}}^i ) \right\} }_{i=1}^{N_{\text{STL}}}$  are generated from ${D_{\text{S}}}$ and ${D_{\text{TL}}}$, based on the proposed Inter-domain Point-CutMix module. The mixed samples ${D}_{\text{STL}}$ serve as the intermediate domain samples to help the model learn domain-invariant features and thus close the inter-domain gap. By training on ${D}_{\text{STL}}$, the knowledge has been transferred from $D_{\text{S}}$ to $D_{\text{TL}}$,  and the obtained model achieves impressive results on the target domain. Therefore, in the second stage, we utilize this model as the teacher detector to learn towards Intra-domain Generalization. Specifically, we use the teacher model to generate pseudo labels $\hat{y}$ for unlabeled target data, which is denoted as ${D_{\text{TU}}}={ \left\{  (x_{\text{TU}}^i, \hat{y}_{\text{TU}}^i)\right\} }_{i=1}^{N_{\text{TU}}}$. Then, the pseudo labeled point cloud ${D_{\text{TU}}}$, together with the real-labeled point cloud $D_{\text{TL}}$, are globally mixed by our Intra-Domain Point-MixUp module to obtain ${D_{\text{TLU}}}={ \left\{  (\tilde{x}_{\text{TLU}}^i, \hat{y}_{\text{TLU}}^i)\right\} }_{i=1}^{N_{\text{TLU}}}$. The mixed samples ${D_{\text{TLU}}}$ are then used to train the student detector. In particular, ${D_{\text{TLU}}}$ tactfully regularizes the pseudo label distribution and thus 
obtain better performance.

\subsection{Inter-domain Point-CutMix}
\label{subsec:cutmix}
% To address the inter-domain discrepancy between source and target domain, an artful solution is to learn domain-invariant features. 
To resolve the inter-domain discrepancy, we propose to mix the LiDAR points from different domains to transfer the data distribution from their original domains to an intermediate domain. This can make the model learn domain-invariant features and integrate the knowledge from both domains. To this end, we present the Point-CutMix module, which is inspired by CutMix~\cite{yun2019cutmix}. Basically, there are three design choices when performing point cloud mixing, \eg, the object level, the region level or the scene level. In this work, we prefer the region level domain mixing. The intuition is that, the global scene-wise features from different domains may be quite different and focusing on the object-level features is yet too strict, making both too difficult to learn. The local region features, however, usually containing similar contexts like the combinations of cars, road planes or buildings, are much easier to learn. Thus it would be more favorable to learn from the locally mixed point cloud samples. 

Specifically, given a point cloud $x_\text{TL}$ from the target domain, we first randomly choose a point in $x_\text{TL}$ as the region center $c_\text{T}$. Then, we randomly select a rectangle region around $c_\text{T}$, based on the bird's eye view (\textit{x-y} plane). After that, we remove all the points outside the selected region and only keep the points inside the region, which is referred to as $P_\text{T}$. $P_\text{T}$ is then used to mix the source point cloud $x_\text{S}$. In particular, we sample another center $c_\text{S}$ in $x_\text{S}$, where $c_\text{S}$  is constrained to have the same range to $c_\text{T}$. Then, we remove the points around $c_\text{S}$ with the same region as that in the target, and use $P_\text{T}$ to inpaint this region. This leads to a new point cloud $\tilde{x}_{STL}$. Similarly, the new label $\tilde{y}_{STL}$ is also obtained by combining the 3D boxes in $P_\text{T}$ and the remained 3D boxes in $x_\text{S}$. Formally, this process can be denoted as:
\begin{align}
    \tilde{x}_\text{STL} &= Concat(M_{x} \odot x_\text{S}^i, (1-M_{x}) \odot x_\text{TL}^j) \\ 
    \tilde{y}_\text{STL} &= Concat(M_{y} \odot y_\text{S}^i, (1-M_{y}) \odot y_\text{TL}^j)
\end{align}
where $M_{x}$, $M_{y}$ denote two binary masks indicating whether this point or label (\ie, 3D box) is inside the selected point cloud region, $\odot$ is the element-wise multiplication operation, and $Concat(\cdot, \cdot)$ denotes the concatenation operation. In this way, we can update the original point cloud distribution to the mixed distribution. The newly generated point clouds contain both source and target local region information, and thus both domain knowledge is reserved. Furthermore, by enforcing the model to learn on the mixed point cloud at each iteration step, we also ensure the data balance of different domains. The Point-CutMix module is applied with a fixed probability during training. After the learning of the Inter-domain Adaptation stage, we can get a domain adaptive model $F_\text{Tea}$, which will be exploited as a teacher model in the next stage.
% Compared with MixUp, we argue that region-wise CutMix can better preserve the local property of original point clouds because 
\subsection{Intra-Domain Point-MixUp}
\label{subsec:mixup}
The main purpose of the second stage is to strengthen the intra-domain generalization capacity to improve the learning on the unlabeled data, which is also known as semi-supervised learning (SSL). Self-training, also called Psuedo-labeling, is a commonly used technique for SSL. Since we have obtained an excellent detector from the first stage, we can utilize it to produce pseudo labels on the unlabeled point cloud. However, there is a potential problem if directly using the pseudo labels, \eg, the existence of inaccurate predictions will inevitably make the distribution of the pseudo labels inconsistent with the real labels. To this end, we propose the Intra-Domain Point-MixUp in this stage to overcome the inconsistent label distribution. Our Point-MixUp is motivated by MixUp~\cite{zhang2017mixup}, aiming to globally mix the point cloud scenes from the labeled and unlabeled sets. Though scene-wise global mixing is not suitable for cross-domain learning, we find that it is extremely beneficial to intra-domain learning, since there is little domain gap between the data from the same domain. By applying the Point-MixUp, the pseudo label can be regularized by the real label, thus improving the label distribution. Moreover, since the unlabeled scale $N_\text{TU}$ is much larger than the labeled scale $N_\text{TL}$, the Point-MixUp also enforces a balanced learning between $D_\text{TL}$ and $D_\text{TU}$.

%What's more, since the amount of labeled and pseudo-labeled target samples varies greatly, it is highly likely that the optimization process will be dominated by pseudo-labeled target samples. The noise in pseudo labels might lead to instability during the training process and performance degradation, so it is of great importance to regularize the training process. Inspired by MixUp~\cite{zhang2017mixup} which was proposed and proved to be a strong regularizer and can improve the generalization capability of the model, we propose Intra-Domain Point-MixUp to mix intra-domain samples at scene wise. 
% Specifically, given two target domain point clouds, we first shuffle all the points. Then we select $\lambda\%$ points and $(100-\lambda)\%$ points of the original point cloud respectively to form a new point cloud. While for the labels, we concatenate all the boxes from both point clouds.

More specifically, given an unlabeled point cloud, we first exploit the teacher model $F_\text{Tea}$ to generate the pseudo labels, \ie, $\hat{y}_{TU} = F_\text{Tea}(x_{TU})$, where $\hat{y}_{TU}$ denotes the pseudo labels. Here, we follow SSL methods such as FixMatch~\cite{FixMatch_2020fixmatch} to utilize a score threshold to filter the pseudo boxes with low confidence. Then, we apply the Point-MixUp on both the data and label levels.
Concretely, given a real-labeled point cloud $(x_{TL}^i, y_{TL}^i )$ and a pseudo-labeled point cloud  $(x_{TU}^i, \hat{y}_{TU}^i )$,  we randomly reserve partial points of $x_{TL}^i$ and $x_{TU}^i$, respectively. This can be implemented by first shuffling the points, and then selecting the top point indices, which are denoted as:
\begin{align}
    \tilde{x}_\text{TLU} &= Concat({P} \odot x_\text{TL}^i, Q \odot x_\text{TU}^j) \\ 
    \hat{y}_\text{TLU} &= Concat(y_\text{TL}^i,  \hat{y}_\text{TU}^j)
\end{align}
where $P$ and $Q$ are two binary masks used for selecting points and are subjected to $\frac{|P=1|}{|P|} + \frac{|Q=1|}{|Q|} = 1$, which ensures the density of the mixed point cloud remains similar as the original ones.

%we first shuffle all the points inside each point cloud1
%\begin{align}
%    x_{TL}^i &= S(x_{TL}^i) \\
%    x_{TU}^j &= S(x_{TU}^j)
%\end{align}
%where $S(\cdot)$ denotes shuffling the points in the given point cloud. Then we sample points from each one
%\begin{align}
%    P &= len(x_{TL}^i)*\lambda\\
%    Q &= len(x_{TU}^j) * (1 - \lambda)
%\end{align}
%where $len(\cdot)$ denotes the total number of points in the given point cloud and $\lambda \in [0, 1]$. Finally, the mixed sample is generated
%\begin{align}
%    \tilde{x}_{TLU} &= Concat(x_{TL}^i[\ :P], x_{TU}^j[\ :Q]) \\
%    \tilde{y}_{TLU} &= Concat(y_{TL}^i, y_{TU}^j)
%\end{align}
% In nature scenes, distant objects often have much fewer points, and our random drops of points on each sample can be deemed as a strong augmentation to simulate the sparse areas. 
Also, before merging the 3D boxes, collision detection will be performed. If there is a collision between the boxes from real and pseudo labels, the pseudo boxes and the near points will be replaced by the real boxes and corresponding points. Afterwards, the mixed samples ${D_{TLU}}={ \left\{  (\tilde{x}_{TLU}^i, \hat{y}_{TLU}^i)\right\} }_{i=1}^{N_{TLU}}$ will be used to train the student detector $F_\text{Stu}$ with a fixed probability, where $F_\text{Stu}$ is initialized by the model in the first stage. We find that the updated label distribution of $\hat{y}_\text{TLU}$ essentially improves the distribution of original pseudo labels $\hat{y}_\text{TU}$ by involving the real labels. By learning on the mixed samples, the model is also encouraged to be more robust on corrupt point cloud samples. 
% which can be formulated as
% \begin{align}
%     x_{gene} &= \{ x_{TL}, x_{TU}, x_{TLU} \}\\
%     y_{gene} &= \{ y_{TL}, y_{TU}, y_{TLU} \}
% \end{align}
% where $x_{gene}$ and $y_{gene}$ denote the training samples and corresponding labels for stage Intra-domain Generalization Learning Stage. Then we minimize the conventional detection loss
% \begin{align}
%     \mathcal{L}_{gene} &= \mathcal{L}_{det}(y_{gene}, \mathcal{F}(x_{gene}))
% \end{align}
% where $\mathcal{F}(\cdot)$ denotes the detector to be trained. 
% \begin{equation}
%     x_t^i = RS(x_t^i)
%     x_u^j = RS(x_u^j)
% \end{equation}

% \begin{equation}
%     \begin{split}
%         \tilde{x}_{intra} = P \odot x_t^i + Q \odot x_t^j \\ 
%         \tilde{y}_{intra} = C(y_s^i, y_u^j)
%     \end{split}
% \end{equation}

% Use \bibliography{yourbibfile} instead or the References section will not appear in your paper
\section{Experiments}

\begin{table*}[h]
\centering
\renewcommand\arraystretch{2}

    \scalebox{0.58}{
\begin{tabular}{c|c|c|c|c|c|c|c|c|c|c} 
\bottomrule
\multirow{3}{*}{\textbf{Methods}} & \multicolumn{10}{c}{\textbf{Semi-supervised Domain Adaptation 3D Detection Performance with Different Target Label Amounts}}                                                                                                                                                                                                                                                                                               \\
                                  & \multicolumn{2}{c|}{\textbf{1\%}}                                    & \multicolumn{2}{c|}{\textbf{5\%}}                                    & \multicolumn{2}{c|}{\textbf{10\%}}                                             & \multicolumn{2}{c|}{\textbf{20\%}}                                             & \multicolumn{2}{c}{\textbf{100\%}}                                                                         \\ 
\cline{2-11}
                                  & \multicolumn{1}{c|}{\textbf{AP} / \textbf{NDS}} & \multicolumn{1}{c|}{\textbf{Closed Gap}}        & \multicolumn{1}{c|}{\textbf{AP} / \textbf{NDS}} & \multicolumn{1}{c|}{\textbf{Closed Gap}}        & \multicolumn{1}{c|}{\textbf{AP} / \textbf{NDS}} & \multicolumn{1}{c|}{\textbf{Closed Gap}}                  & \multicolumn{1}{c|}{\textbf{AP} / \textbf{NDS}} & \multicolumn{1}{c|}{\textbf{Closed Gap}}                  & \multicolumn{1}{c|}{\textbf{AP} / \textbf{NDS}} & \textbf{Closed Gap}          \\ 
\hline
\textbf{Source Only}              & \multicolumn{1}{c|}{42.6 / 50.3}                             & \multicolumn{1}{c|}{+0\% / +0\%}                & \multicolumn{1}{c|}{42.6 / 50.3}                             & \multicolumn{1}{c|}{+0\% / +0\%}                & \multicolumn{1}{c|}{42.6 / 50.3}                             & \multicolumn{1}{c|}{+0\% / +0\%}                          & \multicolumn{1}{c|}{42.6 / 50.3}                             & \multicolumn{1}{c|}{+0\% / +0\%}                          & \multicolumn{1}{c|}{42.6 / 50.3}                                                        & +0\% / +0\%                           \\ 
\hline
\textbf{ST3D-UDA}                 & \multicolumn{1}{c|}{43.7 / 50.2}                             & \multicolumn{1}{c|}{+3.07\% / -0.51\%}          & \multicolumn{1}{c|}{43.7 / 50.2}                             & \multicolumn{1}{c|}{+3.07\% / -0.51\%}          & \multicolumn{1}{c|}{43.7 / 50.2}                             & \multicolumn{1}{c|}{+3.07\% / -0.51\%}                    & \multicolumn{1}{c|}{43.7 / 50.2}                             & \multicolumn{1}{c|}{+3.07\% / -0.51\%}                    & \multicolumn{1}{c|}{43.7 / 50.2}                                                        & +3.07\% / -0.51\%                     \\ 
\hline\hline
\textbf{Labeled Target}           & \multicolumn{1}{c|}{37.2 / 38.1}                             & \multicolumn{1}{c|}{-15.1\% / -62.2\%}          & \multicolumn{1}{c|}{61.0 / 53.2}                             & \multicolumn{1}{c|}{+51.4\% / +14.8\%}          & \multicolumn{1}{c|}{65.6 / 58.2}                             & \multicolumn{1}{c|}{+64.2\% / +40.3\%}                    & \multicolumn{1}{c|}{70.2 / 63.0}                             & \multicolumn{1}{c|}{+77.1\% / +64.8\%}                    & \multicolumn{1}{c|}{78.4 / 69.9}                                                        & +100\% / +100\%                       \\ 
\hline
\textbf{Co-training}              & \multicolumn{1}{c|}{51.4 / 54.6}                             & \multicolumn{1}{c|}{+24.6\% / +21.9\%}          & \multicolumn{1}{c|}{57.7 / 58.0}                             & \multicolumn{1}{c|}{+42.2\% / +39.3\%}          & \multicolumn{1}{c|}{59.4 / 58.9}                             & \multicolumn{1}{c|}{+46.9\% / +43.9\%}                    & \multicolumn{1}{c|}{63.1 / 61.1}                             & \multicolumn{1}{c|}{+57.3\% / +55.1\%}                    & \multicolumn{1}{c|}{66.5 / 63.5}                                                        & +66.8\% / +67.3\%                     \\ 
\hline
\textbf{ST3D-SSDA}                         & \multicolumn{1}{c|}{55.2 / 55.8}                    & \multicolumn{1}{c|}{+35.2\% / +28.1\%} & \multicolumn{1}{c|}{63.9 / 60.8}                    & \multicolumn{1}{c|}{+59.5\% / +53.6\%} & \multicolumn{1}{c|}{68.2 / 63.2}                    & \multicolumn{1}{c|}{+71.5\% / +65.8\%}           & \multicolumn{1}{c|}{71.4 / 65.3}                    & \multicolumn{1}{c|}{+80.4\% / +76.5\%}           & \multicolumn{1}{c|}{\color[rgb]{0,0.4,0}{79.7 / 71.3}}                                               & {\color[rgb]{0,0.4,0}+103.6\% / +107.1\%}  \\ 
\hline
\textbf{Ours}                     & \multicolumn{1}{c|}{\textbf{73.4 / 67.1}}                    & \multicolumn{1}{c|}{\textbf{+86.0\% / +85.7\%}} & \multicolumn{1}{c|}{\textbf{76.2 / 68.8}}                    & \multicolumn{1}{c|}{\textbf{+93.9\% / +94.4\%}} & \multicolumn{1}{c|}{\color[rgb]{0,0.4,0}{\textbf{78.8 / 70.9}}}                    & \multicolumn{1}{c|}{\color[rgb]{0,0.4,0}{\textbf{+101.1\% / +105.1\%}}} & \multicolumn{1}{c|}{\color[rgb]{0,0.4,0}{\textbf{78.9 / 71.0}}}                    & \multicolumn{1}{c|}{\color[rgb]{0,0.4,0}{\textbf{+101.4\% / +105.6\%}}} & \multicolumn{1}{c|}{\color[rgb]{0,0.4,0}{\textbf{79.8 / 71.8}}}                                               & \textbf{\color[rgb]{0,0.4,0}{+103.9\% / 109.7\%}}  \\ 
\hline
\textbf{\color{gray}Oracle}               & \multicolumn{1}{c|}{\color{gray}78.4 / 69.9}                         & \multicolumn{1}{c|}{\color{gray}+100\% / +100\%}        & \multicolumn{1}{c|}{\color{gray}78.4 / 69.9}                         & \multicolumn{1}{c|}{\color{gray}+100\% / +100\%}        & \multicolumn{1}{c|}{\color{gray}78.4 / 69.9}                         & \multicolumn{1}{c|}{\color{gray}+100\% / +100\%}                  & \multicolumn{1}{c|}{\color{gray}78.4 / 69.9}                         & \multicolumn{1}{c|}{\color{gray} +100\% / +100\%}                 & \multicolumn{1}{c|}{\color{gray}78.4 / 69.9}                                                    & {\color{gray} +100\% / +100\%}                  \\
\bottomrule
\end{tabular}
}
% \vspace{-1mm}
\caption{\small \textbf{Domain adaptation performance with different amounts of target labels.} We report AP, NDS and their corresponding Closed Gap on the car category for all methods. The best adaptation performance is indicated by \textbf{bold} and \textit{Oracle} is shown in {\color{gray}gray}. We also indicate the performance that surpasses \textit{Oracle} with {\color[rgb]{0,0.4,0} {\textbf{green}}} color.}
\label{table1}
\end{table*}

\begin{table*}[t]
    \centering
    \renewcommand\arraystretch{1.5}
    \resizebox{0.8\textwidth}{!}
    {
    \begin{tabular}{c|ccccc|c}
\bottomrule
    & \multicolumn{5}{c|}{\textbf{Components}}                                                                                                                                                                   & \multirow{2}{*}{\textbf{AP} / \textbf{NDS}} \\ \cline{1-6}
    % & \multicolumn{1}{c}{\textbf{Target Supervised}}         & \multicolumn{1}{c}{\textbf{Source Supervised}} & \multicolumn{1}{c|}{\textbf{Inter-domain CutMix}} & \multicolumn{1}{c|}{\textbf{Naive SSL Learning}} & \textbf{Intra-Domain MixUp} &                           \\ \hline
    & \multicolumn{1}{c|}{\textbf{Target Supervised}}         & \multicolumn{1}{c|}{\textbf{Source Supervised}} & \multicolumn{1}{c|}{\textbf{Inter-domain CutMix}} & \multicolumn{1}{c|}{\textbf{Naive SSL Learning}} & \textbf{Intra-Domain MixUp} &                                 \\ \hline
(a) & \multicolumn{1}{c|}{\checkmark} & \multicolumn{1}{c|}{}                  & \multicolumn{1}{c|}{}                    & \multicolumn{1}{c|}{}                   &                    & 61.0 / 53.2                      \\ \hline
(b) & \multicolumn{1}{c|}{\checkmark}                          & \multicolumn{1}{c|}{\checkmark}                  & \multicolumn{1}{c|}{}                    & \multicolumn{1}{c|}{}                   &                    & 57.7 / 58.0                     \\ \hline
(c) & \multicolumn{1}{c|}{\checkmark}                          & \multicolumn{1}{c|}{\checkmark}                  & \multicolumn{1}{c|}{\checkmark}                    & \multicolumn{1}{c|}{}                   &                    & 74.5 / 67.8                     \\ \hline \hline
(d) & \multicolumn{1}{c|}{\checkmark}                          & \multicolumn{1}{c|}{\checkmark}                  & \multicolumn{1}{c|}{\checkmark}                    & \multicolumn{1}{c|}{\checkmark}                   &                    & 73.8 / 67.7                     \\ \hline 
(e) & \multicolumn{1}{c|}{\checkmark}                          & \multicolumn{1}{c|}{\checkmark}                  & \multicolumn{1}{c|}{\checkmark}                    & \multicolumn{1}{c|}{\checkmark}                   & {\checkmark}                   & 76.2 / 68.8                    \\ \hline
(f) & \multicolumn{1}{c|}{\checkmark}                          & \multicolumn{1}{c|}{}                  & \multicolumn{1}{c|}{}                    & \multicolumn{1}{c|}{\checkmark}                   & {\checkmark}                   & 65.4 / 58.2                    \\ \hline
\end{tabular}
}
% \vspace{-1mm}
    \caption{\small \textbf{Ablation studies for each component of~\ourmodel.} These experimental results are reported based on 5\% target label.}
    \label{table2}
    % \vspace{-3mm}
\end{table*}

\subsection{Experiment Setup} 
\textbf{Datasets.} Our experiments are conducted on two widely used datasets: Waymo\cite{WAYMO_2020Sun} with 64-beam LiDAR and nuScenes\cite{caesar2019nuscenes} with 32-beam LiDAR. We adapt from Waymo to nuScenes, \ie, 100\% Waymo annotations together with partial nuScenes annotations are used. In particular, we uniformly downsample the nuScenes training samples into 1\%, 5\%, 10\%, 20\% and 100\% (resulting in 282, 1407, 2813, 5626 and 28130 frames), and the rest of the samples remain unlabeled. \\

\noindent\textbf{Comparison Methods}. We compare \ourmodel~with six methods: (\textit{\textbf{i}}) \textbf{Source Only} indicates that we directly evaluate the model on target after training on source. (\textit{\textbf{ii}}) \textbf{ST3D-UDA} indicates the ST3D under UDA setting where no real target labels are available. (\textit{\textbf{iii}}) \textbf{Labeled Targets} indicates that only partial target labels (\ie, 1\%, 5\%, 10\%, 20\% or 100\%) are used to train the model with fully-supervised learning.
% leveraging only a few real target labels to train the model in a conventional supervised fashion. 
(\textit{\textbf{iv}}) \textbf{Co-training} represents that we train labeled source and target data jointly. (\textit{\textbf{v}}) \textbf{ST3D-SSDA} is the extension of \textbf{ST3D-UDA} with extra supervision on limited real labeled target data. (\textit{\textbf{vi}}) \textbf{Oracle} indicates the fully supervised model trained on target domain. \\

\noindent\textbf{Evaluation Metric.} Following  nuScenes, we also select commonly used average precision (AP) and the specific metric nuScenes detection score (NDS) in nuScenes as our evaluation metric on the car category, which is also named Vehicle in Waymo. We follow the official nuScenes protocol to average over matching thresholds of $\mathbb{D}$=\{0.5, 1, 2, 4\} meters on the car category. NDS is proposed to consolidate the different error types into a scalar score between 0 and 1, the higher the better. What's more, following ST3D, we also report closed gap which is defined as $\frac{AP_{\text{model}} - AP_{\text{source~only}}}{AP_{\text{oracle}} - AP_{\text{source~only}}} \times 100\%$ to show how much the performance gap is closed. \\

\noindent\textbf{Implementation Details.} All the methods are implemented based on an advanced 3D detector, CenterPoint\cite{CENTERPOINT_2021}. The learning schedule follows the popular codebase OpenPCDet\cite{od2020openpcdet}, where the training epochs are set to 20 for both stages. 
% The probability of Inter-domain Point-CutMix and Intra-Domain Point-MixUp is set to be 0.5 empirically. 
For both Inter-domain Point-CutMix and Intra-Domain Point-MixUp, there is a probability to decide whether to utilize the corresponding technique. We conduct grid-search experiments and set them as 0.5. 
The labeled target samples are downsampled in a frame-level. For example, 1\% labeled target data means that only 1\% of the total samples are available and the GT-database for GT-sampling augmentation is also determined accordingly. 
The detection range is set to [-54.0, 54.0]\textit{m} for \textit{X} and \textit{Y} axes, and [-5.0, 4.8]\textit{m} for \textit{Z} axis. We set the voxel size to [0.075, 0.075, 0.2]. As there is a difference in the range of intensity between Waymo and nuScenes, we normalize it to $0\sim1$ for both datasets. For augmentation techniques, we adopt widely used random world flip, random world rotation and random world scaling for both learning stages.

\subsection{Performance Comparison} 
We perform domain adaptation experiments from Waymo to nuScenes with different amounts of nuScenes label . The 3D detection performance of different methods is shown in Table \ref{table1}. As can be seen, our method achieves the best performance and surpasses all the other methods by a large margin with all settings. Specifically, with only 1\% labeled target data, we improve the performance of Source Only and ST3D-UDA by around 30\% AP and 17\% NDS. We can still gain noticeable performance improvements of 18.2\% AP and 11.3\% NDS compared with ST3D-SSDA. Besides, with the increasing number of labeled target data, our method continuously produces better results. Compared with other methods, our method can greatly close the gap between \textit{Oracle} and Source Only. With only 5\% labeled target data, we achieve 94\% Closed Gap. What's more, it is worth noting that with only 10\% labeled target data, we even {surpass} \textit{Oracle} in terms of AP and NDS (\ie, 78.8\% \textit{vs} 78.4\% AP and 70.9\% \textit{vs} 69.9\% NDS). To sum up, our method can save about 90\% annotation cost in cross-domain LiDAR-based 3D object detection.

\begin{figure}[tbp]
    \centering %%
    % \subfigure[\scriptsize Source]{\includegraphics[width=0.15\textwidth]{v0/figs/waymo_75_w_box.png}}
    % \subfigure[\scriptsize Real-labeled Target]{\includegraphics[width=0.15\textwidth]{v0/new_nus_75_w_box.png}}
    % \subfigure[\scriptsize CutMixed Sample]{\includegraphics[width=0.15\textwidth]{v0/new_cutmix_75_w_box.png}}
    \subfigure[\footnotesize Qualitative results of Inter-domain Point-CutMix.]{\includegraphics[width=0.45\textwidth]{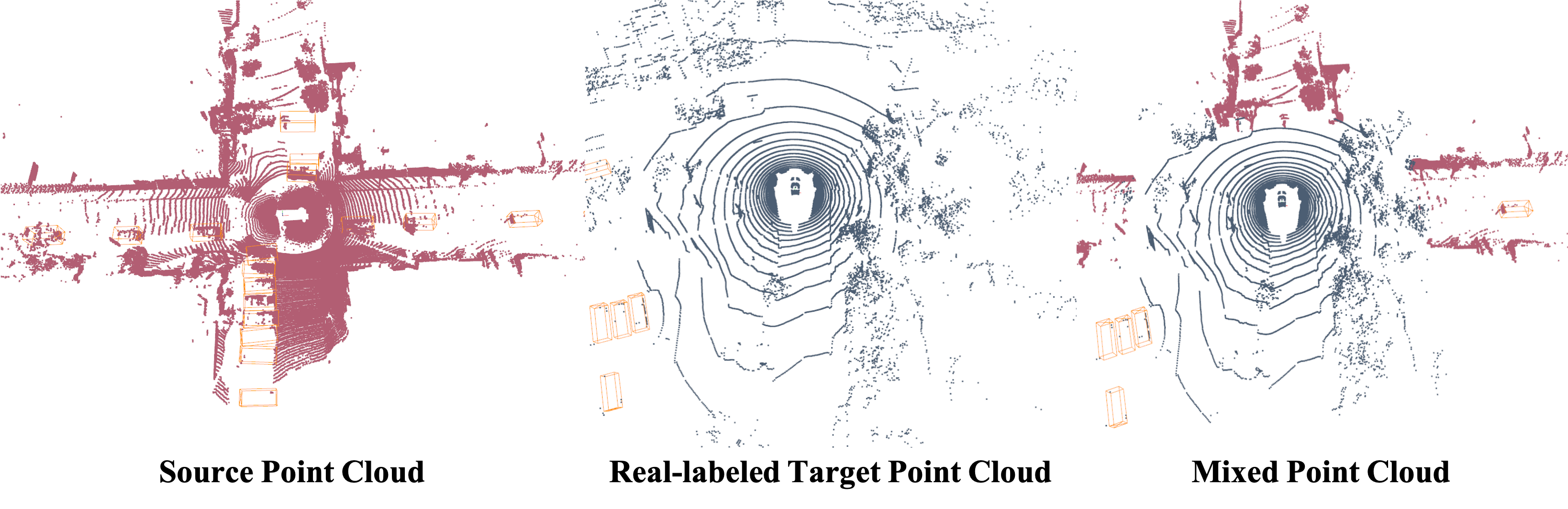}}
    % \label{fig:cutmix}
    \\
    \subfigure[\footnotesize Qualitative results of Intra-domain Point-MixUp.]{\includegraphics[width=0.45\textwidth]{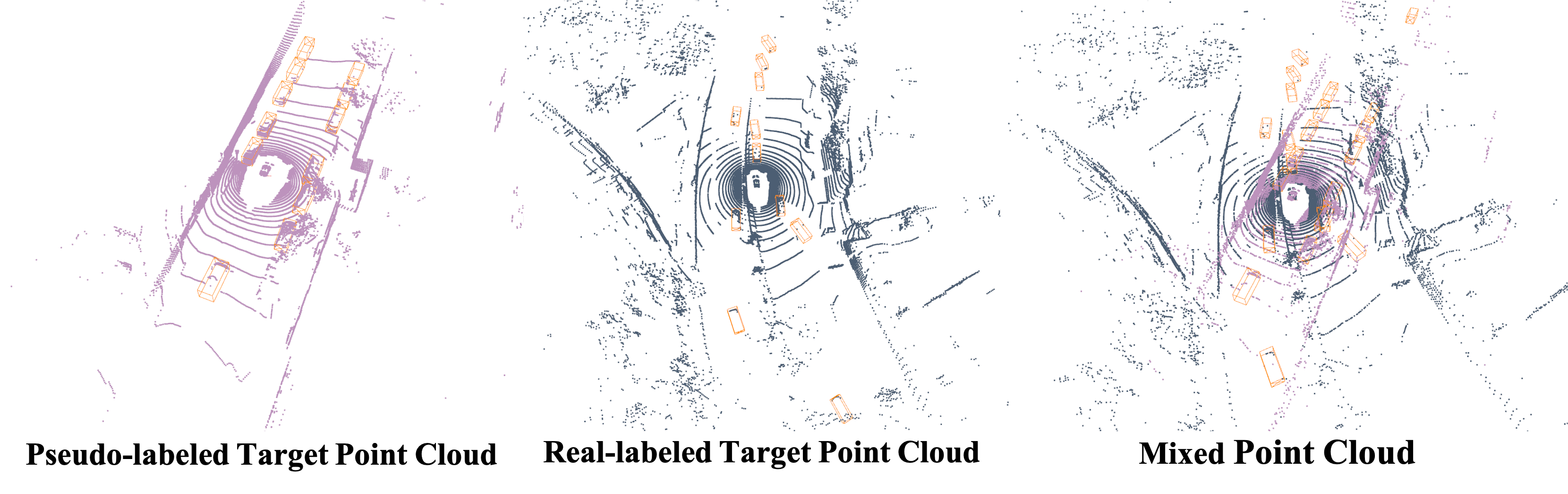}}
    % \vspace{-3mm}
    % \caption{Qualitative results of Inter-domain Point-CuMix.}
    % \includegraphics[width=0.5\linewidth]{v0/figs/cutmix_75_mixup_2_w_box.png}
    % \caption{\small Qualitative results of Inter-domain Point-CutMix(\textit{top}) and Intra-domain Point-Mixup(\textit{bottom}). In the top row, from left to the right are original Waymo sample, original real-labeled nuScens sample and mixed sample after Inter-domain Point-CutMix. In the bottom row, from left to the right are original pseudo-labeled nuScenes sample, original real-labeled nuScenes sample and mixed sample after Intra-domain Point-MixUp.}
    % \label{fig:mixup}
    % \vspace{-3mm}
    \caption{\small \textbf{Qualitative results} of different types of mixed samples.}
    \label{visualization}
    % \vspace{-3mm}
\end{figure}

\subsection{Ablation Studies}
\subsubsection{Effectiveness of Each Component.}
In this section, we conduct ablation experiments to validate the effectiveness of each individual component of our proposed method, which is shown in Table \ref{table2}. All the experiments are conducted with 5\% labeled target data. From the comparison of (a) and (b), we can see that simply training the combined data of labeled source and target data will not yield performance gains. Instead, it causes performance degradation. This proves that the giant domain gap between source and target could hinder the model in learning reasonable features and the large-scale source data also dominates the learning.
By contrast, after applying Inter-domain Point-CutMix, an obvious improvement can be seen from the comparison between (b) and (c), \ie, AP and NDS are improved by nearly 20\% and 10\%, respectively. It turns out that the Inter-domain Point-CutMix can help to learn domain-invariant features that are helpful in improving the 3D detection performance in the target domain. \\
Afterwards, we investigate the effectiveness by involving the unlabeled target data. In (d), we first utilize a naive semi-supervised learning method that simply trains the model using both the real- and pseudo-labeled target samples. However, this obtains nearly no gains compared with (c). While after we apply the Intra-Domain Point-MixUp, the performance improves from 73.8\% to 76.2\% in terms of AP and from 67.7\% to 68.8\% in terms of NDS. It clearly suggests the plausibility of the Intra-Domain Point-MixUp in regularizing the learning. From (e), we can further tell that learning from source domain contributes most to the final performance. Transferring knowledge from source domain to target domain largely improves the performance on the target domain.\\

\subsubsection{Qualitative Results}
To better understand the mixed point cloud samples after the Inter-domain Point-CutMix and Intra-domain Point-MixUp modules. We  give some examples of them in Figure \ref{visualization}. As can be seen in Figure~\ref{visualization} (a), the Inter-domain Point-CutMix module produces new scenes that resemble the nature scenes due to the constraining of the range of the mixed regions. Further, the Intra-domain Point-MixUp module yields sparser point clouds from the perspective of each involved sample, which also effectively regularizes the point cloud distribution. 

% \begin{figure}[htbp]
%     \centering 
%     \subfigure[\scriptsize Pseudo-labeled Target]{\includegraphics[width=0.15\textwidth]{v0/figs/tu_2_w_box.png}}
%     \subfigure[\scriptsize Real-labeled Target]{\includegraphics[width=0.15\textwidth]{v0/figs/new_tl_2_w_box.png}}
%     \subfigure[\scriptsize MixUped Sample]{\includegraphics[width=0.15\textwidth]{v0/figs/new_mixup_2_w_box.png}}
%     \vspace{-3mm}
%     \caption{Qualitative results of Intra-domain Point-MixUp.}
%     % \includegraphics[width=0.5\linewidth]{v0/figs/cutmix_75_mixup_2_w_box.png}
%     % \caption{\small Qualitative results of Inter-domain Point-CutMix(\textit{top}) and Intra-domain Point-Mixup(\textit{bottom}). In the top row, from left to the right are original Waymo sample, original real-labeled nuScens sample and mixed sample after Inter-domain Point-CutMix. In the bottom row, from left to the right are original pseudo-labeled nuScenes sample, original real-labeled nuScens sample and mixed sample after Intra-domain Point-MixUp.}
%     \label{fig:mixup}
%     \vspace{-5mm}
% \end{figure}

\subsubsection{Probability of Inter-domain Point-CutMix. } 
\begin{figure}[t]
    \centering
    \includegraphics[width=\linewidth]{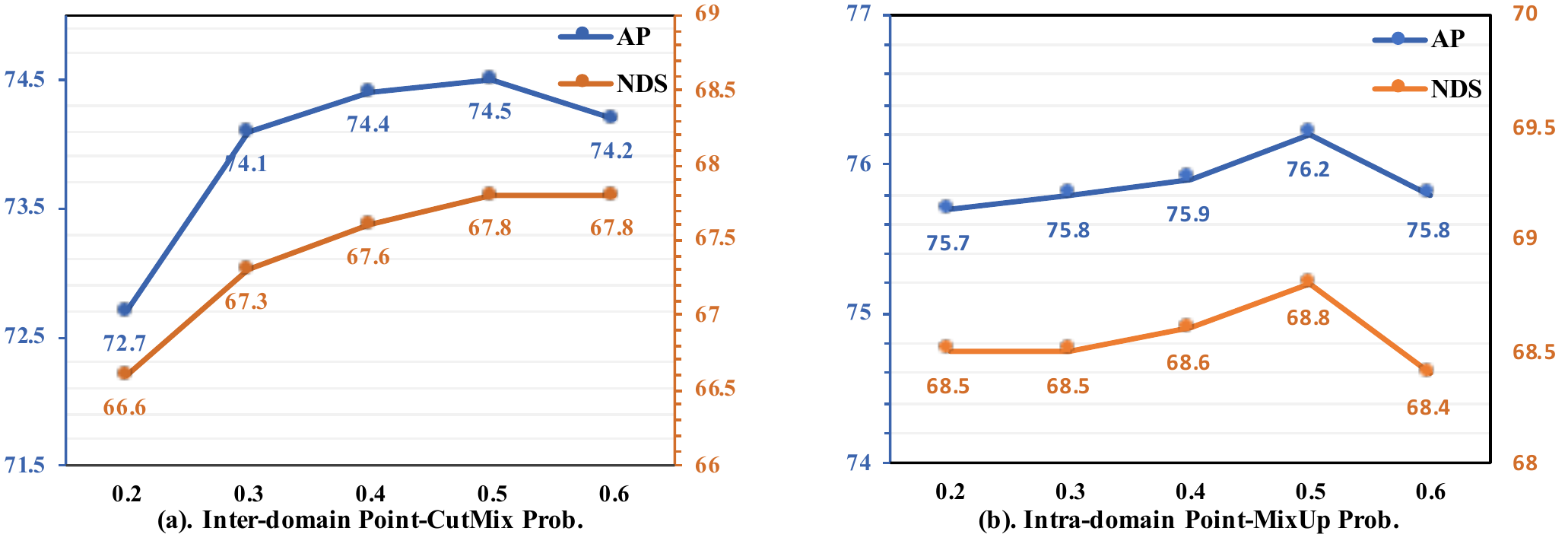}
    % \vspace{-7mm}
    \caption{\small \textbf{Ablation studies} for mixing probability. (a) and (b) indicate the impact of Inter-domain Point-CutMix probability and Intra-domain Point-MixUp probability, respectively.}
    \label{fig:figure3}
    % \vspace{-2mm}
\end{figure}

In this section, we validate the effectiveness of the Inter-domain Point-CutMix module by adjusting its probability. The results are shown in Figure \ref{fig:figure3} (a). As seen, the probability of applying Inter-domain Point-CutMix has an obvious impact on the final AP and NDS performance. In general, the AP and NDS continue to get better as the probability increases, and the best performance is obtained when the probability is set to 0.5. It is interesting that even with a probability of 0.2, the performance is still better than the model without the Point-CutMix module, \ie, the Co-training model (57.7\% AP and 58.0\% NDS) that jointly trains the labeled source and target data. This further proves the necessity of our Inter-domain Point-CutMix module.
% follow the same trend that with the probability increasing the performance gets better and begins to drop after making it to the top at 0.5 , which is 74 .5\% in AP and 67.8\% in NDS. 
% What's more, even , the performance is much better than jointly training the labeled source and target data, which  at region wise.

%----------- Figure 3 ----------- %
% \begin{figure}[t]
% \centering     %%% not \center
% \begin{subfigure}
% \centering
% \includegraphics[width=0.47\linewidth]{camera_ready/figs/thresh_cutmix.pdf}
% \end{subfigure}
% \begin{subfigure}
% \centering
% \includegraphics[width=0.47\linewidth]{camera_ready/figs/thresh_mixup.pdf}
% \end{subfigure}

% \vspace{-4mm}
% \caption{\small \textbf{Ablation studies} for mixing probability. (a) and (b) indicates the impact of Inter-domain Point-CutMix probability and Intra-domain Point-MixUp probability respectively.}
% \label{fig:figure3}
% \vspace{-2mm}
% \end{figure}

\subsubsection{Probability of Intra-domain Point-MixUp. } In this section, we conduct experiments to validate the impact of the probability to apply Intra-Domain Point-MixUp. As shown in Figure \ref{fig:figure3} (b), a similar trend to (a) can be seen. As the probability increases, the performance keeps improving and reaches the peak with 76.2\% in AP and 68.8\% in NDS at the probability of 0.5. Higher probability doesn't give better results. All the performances are better than the model without the Point-MixUp (73.8\% AP and 67.7\% NDS). This also indicates the robustness and generalization ability of our model.

\section{Conclusion}
In this paper, we presented a new task, semi-supervised domain adaptation (SSDA) in the context of 3D object detection, as well as a novel framework \ourmodel. We decoupled the learning of SSDA-based 3D object detection into two stages, which are the Inter-domain Adaptation stage and the Intra-domain Generalization stage. The purposes are to address the cross-domain discrepancy and improve semi-supervised learning, respectively. During the Inter-domain Adaptation stage, we proposed a Point-CutMix module to construct mixed point cloud samples to learn domain-invariant features. Then, in the Intra-domain Generalization stage, a Point-MixUp module was advocated to regularize the pseudo label distribution. Extensive experiments prove the effectiveness of \ourmodel, \eg, on Waymo$\rightarrow$nuScenes, and we significantly outperform previous methods. Moreover, with 10\% labeled target data,  \ourmodel~achieves superior performance to the fully-supervised \textit{Oracle} model.

\section*{Acknowledgements}
This work was supported in part by the FDCT grant SKL-IOTSC(UM)-2021-2023, and the Start-up Research Grant (SRG) of University of Macau (SRG2022-00023-IOTSC). We would like to thank Inceptio for their infra supports. 
\bibliography{1879_camera_ready.bib}
\end{document}